\pdfoutput=1

\documentclass[11pt]{article}

\usepackage[]{ACL2023}

\usepackage{times}
\usepackage{latexsym}
\usepackage{graphicx}
\usepackage[T1]{fontenc}

\usepackage[utf8]{inputenc}

\usepackage{microtype}

\usepackage{soul}
\usepackage{color}
\usepackage{inconsolata}

\usepackage{xurl}

\newcommand{\nop}[1]{}

\newcommand{\spidert}{Spider$_T$}
\newcommand{\spiderd}{Spider$_D$}
\newcommand{\cgsubt}{CG-SUB$_T$}
\newcommand{\cgsubd}{CG-SUB$_D$}
\newcommand{\cgappt}{CG-APP$_T$}
\newcommand{\cgappd}{CG-APP$_D$}

\usepackage{adjustbox}
\newcommand{\bftab}{\fontseries{b}\selectfont}
\usepackage{booktabs}
\usepackage{multirow}
\usepackage{multicol}
\usepackage{tabularx}
\usepackage{amsfonts}
\usepackage{comment}

\newcommand{\ctext}[3][RGB]{%
  \begingroup
  \definecolor{hlcolor}{#1}{#2}\sethlcolor{hlcolor}%
  \hl{#3}%
  \endgroup
}

\title{Improving Generalization in Language Model-Based Text-to-SQL Semantic Parsing: Two Simple Semantic Boundary-Based Techniques}

\author{Daking Rai\textsuperscript{1}, Bailin Wang\textsuperscript{2}, Yilun Zhou\textsuperscript{2}, Ziyu Yao\textsuperscript{1}\\
        \textsuperscript{1}George Mason University, \textsuperscript{2}MIT\\
        \textsuperscript{1}{\tt \{drai2, ziyuyao\}@gmu.edu}, \textsuperscript{2}{\tt\{bailinw, yilun\}@mit.edu}\\
}

\begin{document}
\maketitle
\begin{abstract}
Compositional and domain generalization present significant challenges in semantic parsing, even for state-of-the-art semantic parsers based on pre-trained language models (LMs). In this study, we empirically investigate improving an LM's generalization in semantic parsing with two simple techniques: at the \emph{token} level, we introduce a token preprocessing method to preserve the semantic boundaries of tokens produced by LM tokenizers; at the \emph{sequence} level, we propose to use special tokens to mark the boundaries of components aligned between input and output. Our experimental results on two text-to-SQL semantic parsing datasets show that our token preprocessing, although simple, can substantially improve the LM performance on both types of generalization, and our component boundary marking method is particularly helpful for compositional generalization.\footnote{{The source code for our implementation is available at \url{https://github.com/Dakingrai/ood-generalization-semantic-boundary-techniques}.}}
\end{abstract}

\section{Introduction}\label{sec:intro}
Pre-trained language models (LMs)\footnote{We use ``LMs'' to refer to a broad set of models that are pre-trained in (masked/autoregressive) language modeling objectives, with encoder-decoder or decoder-only architecture.} such as T5 \cite{raffel2020exploring} have now been more and more widely adopted for semantic parsing due to their promising performance and straightforward architectures \cite{shaw-etal-2021-compositional, scholak-etal-2021-picard, yin-etal-2021-compositional, qi2022rasat, xie2022unifiedskg, qiu2021improving}. However, recent work revealed that these LMs still struggle to generalize on out-of-distribution (OOD) samples \citep{pmlr-v80-lake18a, keysers2019measuring, shaw-etal-2021-compositional, qiu2022evaluating}. 
For example, if a parser has learned ``how many heads are in the department'' and ``how many people are older than 56'', it is expected to generalize to ``how many heads of the departments are older than 56''.

Generalizing to such novel component compositions is known as {\it compositional generalization}. Additionally, generalizing to new domains (e.g., from ``entertainment'' to ``flight'') is referred to as {\it domain generalization}.

\begin{table}[!t]
    \centering
    \resizebox{\columnwidth}{!}{
    \begin{tabular}{p{1cm}p{8cm}}\toprule
    \multicolumn{2}{p{9cm}}{\textit{Token Preprocessing (applied to database schema)}} \\
        \textbf{Before}: & \texttt{
 \ctext[RGB]{255,187,120}{department}\ctext[RGB]{152,223,138}{\_}\ctext[RGB]{255,187,120}{management} \ctext[RGB]{152,223,138}{|} \ctext[RGB]{255,187,120}{department} \ctext[RGB]{152,223,138}{:}\ctext[RGB]{255,187,120}{|} \ctext[RGB]{152,223,138}{i}\ctext[RGB]{255,187,120}{d} \ctext[RGB]{152,223,138}{,} \ctext[RGB]{255,187,120}{budget}\ctext[RGB]{152,223,138}{\_}\ctext[RGB]{255,187,120}{in}\ctext[RGB]{152,223,138}{\_}\ctext[RGB]{255,187,120}{b}\ctext[RGB]{152,223,138}{illion}\ctext[RGB]{255,187,120}{s} \ctext[RGB]{152,223,138}{,} \ctext[RGB]{255,187,120}{num}\ctext[RGB]{152,223,138}{\_}\ctext[RGB]{255,187,120}{e}\ctext[RGB]{152,223,138}{m}\ctext[RGB]{255,187,120}{plo}\ctext[RGB]{152,223,138}{y}\ctext[RGB]{255,187,120}{e}\ctext[RGB]{152,223,138}{e}\ctext[RGB]{255,187,120}{s}}\\
        \textbf{After}: & \texttt{
 \ctext[RGB]{255,187,120}{department}\ctext[RGB]{152,223,138}{\_}\ctext[RGB]{255,187,120}{management} \ctext[RGB]{152,223,138}{|} \ctext[RGB]{255,187,120}{department} \ctext[RGB]{152,223,138}{:}\ctext[RGB]{255,187,120}{|} \ctext[RGB]{152,223,138}{i}\ctext[RGB]{255,187,120}{d} \ctext[RGB]{152,223,138}{,} \ctext[RGB]{255,187,120}{budget} \ctext[RGB]{152,223,138}{\_} \ctext[RGB]{255,187,120}{in} \ctext[RGB]{152,223,138}{\_} \ctext[RGB]{255,187,120}{billion}\ctext[RGB]{152,223,138}{s} \ctext[RGB]{255,187,120}{,} \ctext[RGB]{152,223,138}{num} \ctext[RGB]{255,187,120}{\_} \ctext[RGB]{152,223,138}{employees}}\\\midrule
    
        \multicolumn{2}{p{9cm}}{\textit{Token Preprocessing (applied to SQL)}} \\
        \textbf{Before}: & \texttt{
\ctext[RGB]{255,187,120}{select} \ctext[RGB]{152,223,138}{a}\ctext[RGB]{255,187,120}{v}\ctext[RGB]{152,223,138}{g} \ctext[RGB]{255,187,120}{(} \ctext[RGB]{152,223,138}{flight}\ctext[RGB]{255,187,120}{.}\ctext[RGB]{152,223,138}{p}\ctext[RGB]{255,187,120}{rice}\ctext[RGB]{152,223,138}{)} \ctext[RGB]{255,187,120}{where} \ctext[RGB]{152,223,138}{flight}\ctext[RGB]{255,187,120}{.}\ctext[RGB]{152,223,138}{o}\ctext[RGB]{255,187,120}{rig}\ctext[RGB]{152,223,138}{in} \ctext[RGB]{255,187,120}{=} \ctext[RGB]{152,223,138}{`}\ctext[RGB]{255,187,120}{New} \ctext[RGB]{152,223,138}{York}\ctext[RGB]{255,187,120}{'}}\\
        \textbf{After}: & \texttt{
\ctext[RGB]{255,187,120}{select} \ctext[RGB]{152,223,138}{average} \ctext[RGB]{255,187,120}{(} \ctext[RGB]{152,223,138}{flight} \ctext[RGB]{255,187,120}{.} \ctext[RGB]{152,223,138}{price} \ctext[RGB]{255,187,120}{)} \ctext[RGB]{152,223,138}{where} \ctext[RGB]{255,187,120}{flight} \ctext[RGB]{152,223,138}{.} \ctext[RGB]{255,187,120}{origin} \ctext[RGB]{152,223,138}{=} \ctext[RGB]{255,187,120}{`}\ctext[RGB]{152,223,138}{New} \ctext[RGB]{255,187,120}{York}\ctext[RGB]{152,223,138}{'}}\\\midrule
        \multicolumn{2}{p{9cm}}{\textit{Component Boundary Marking (applied to NL input and SQL output)}}\\
        \textbf{Before}: & How many heads of the departments are older than 56 ? \\
        & \texttt{select count (head.*) where head.age > 56}\\
        \textbf{After}: & \texttt{\textcolor[RGB]{255,127,14}{[sep0]}} How many heads of the departments \texttt{\textcolor[RGB]{255,127,14}{[/sep0]}} \texttt{\textcolor[RGB]{255,127,14}{[sep1]}} are older than 56 ? \texttt{\textcolor[RGB]{255,127,14}{[/sep1]}}\\
        & \texttt{\textcolor[RGB]{255,127,14}{[sep0]} select count (head.*) \textcolor[RGB]{255,127,14}{[/sep0]} \textcolor[RGB]{255,127,14}{[sep1]} where head.age > 56 \textcolor[RGB]{255,127,14}{[/sep1]}} \\\bottomrule
    \end{tabular}
    }
    \caption{Our proposed techniques. Top: we preprocess the text such that its T5 tokenization aligns with word semantics. Coloring indicates tokenization; for example, ``avg'' is converted into three tokens of ``a'', ``v'' and ``g''. Bottom: we add separator tokens to mark the boundaries of aligned semantic components in the input and output.}
    \label{tab:overview}
\end{table}

In this paper, we investigate these two types of generalization of LMs in text-to-SQL semantic parsing, i.e., given a natural language (NL) input and the database schema, producing a SQL query that can be executed against the database for desired output. We conduct experiments using the cross-database Spider benchmark \cite{yu-etal-2018-spider} and its derivation Spider-CG~\cite{gan-etal-2022-measuring}.
Compared with existing benchmarks~\cite{keysers2019measuring, pmlr-v80-lake18a}, this task setting is both more realistic (e.g., containing larger language variations) and more challenging (e.g., requiring grounding to the database context).

Although previous work tackling the two types of generalization all requires non-trivial engineering effort (see Section~\ref{sec:related_work}), in this work, we present two simple yet effective techniques, which are extremely easy to implement with LMs (Table~\ref{tab:overview}). Our techniques improve the generalization of LMs by preserving the \emph{semantic boundaries} at the token and the sequence levels. 
At the token level, our first technique rewrites the inputs to handle naming conventions in database schemas and SQL queries such that a pre-trained LM tokenizer can split them into semantically meaningful tokens.

At the sequence level, our second technique introduces special tokens to mark the semantic boundaries (e.g., phrases) aligned between the source NL and the target SQL. These special tokens implicitly help the LM-based parser build more precise input-output correspondences that are crucial for compositional generalization.

On five evaluation sets, the experimental results based on T5-base show that, albeit simple, our token-level technique dramatically improves both types of LM generalization, and our sequence-level technique is particularly helpful for compositional generalization. Combining them together leads to further improvements. Our additional experiments further demonstrate the generalizability of our approaches (e.g., to text-to-LISP expression parsing \cite{SMDataflow2020}).
\section{Related Work}\label{sec:related_work}

\paragraph{Text-to-SQL Semantic Parsing.} This task has received considerate attention since the creation of the WikiSQL \cite{zhong2017seq2sql} and Spider \cite{yu-etal-2018-spider} datasets. While a large amount of existing work designed specialized architectures for this task \cite{yu-etal-2018-syntaxsqlnet, zhang-etal-2019-editing, wang-etal-2020-rat, lin-etal-2020-bridging}, there has been a trend of directly fine-tuning pre-trained sequence-to-sequence models as semantic parsers \cite{shaw-etal-2021-compositional, scholak-etal-2021-picard, xie2022unifiedskg, qi2022rasat}. Our work follows the same line and proposed approaches to further improve the LM performance. On the other hand, \citet{guo-etal-2019-towards, gan-etal-2021-natural-sql, herzig2021unlocking} showed that simplifying the SQL representation in a way that the new representation can semantically better align with the NL can dramatically improve the parsing performance. In our work, we follow the NatSQL representation \cite{gan-etal-2021-natural-sql} as it has better alignments with the NL.

\vspace{1mm}
\noindent \textbf{Injecting Priors into Semantic Parsers.} 
Our two techniques can be viewed as injecting human prior knowledge into neural models for better generalization, which has been one of the major research efforts on improving domain and 
compositional generalization.
The key consideration to be taken when injecting priors is the trade-off between the form and the generalizability.
Strong priors in the form of specialized model architectures~\citep{shaw-etal-2021-compositional,herzig-berant-2021-span,bailin2021structured} are either too expensive or not applicable across domains. Weaker priors in terms of specialized training algorithms~\citep{yin-etal-2021-compositional,conklin-etal-2021-meta} are more general, but often weaker in performance compared to other lines of methods. Our work is in the spirit of the third line on the use of data augmentation~\citep{andreas-2020-good,akyurek2020learning,qiu-etal-2022-improving}. However, instead of synthesizing new data from scratch, we ``annotate'' the data with semantic boundary markers, which is not only much simpler but also brings better performance.
The final line of work~\citep{qiu2022evaluating,levy2022diverse} is based on the learning capacities in the context of large LMs, which is out of the scope of this work.

\section{Methods}

\subsection{Token Preprocessing}

\begin{table}[!htb]
    \centering
    \resizebox{\columnwidth}{!}{
    \begin{tabular}{ll}\toprule
    Before preprocessing & After preprocessing\\\midrule
    \multicolumn{2}{l}{\textit{Snake case in schema items (add space)}}\\
        \texttt{\ctext[RGB]{255,187,120}{booking}\ctext[RGB]{152,223,138}{\_}\ctext[RGB]{255,187,120}{stat}\ctext[RGB]{152,223,138}{us}\ctext[RGB]{255,187,120}{\_}\ctext[RGB]{152,223,138}{code}} & \texttt{\ctext[RGB]{255,187,120}{booking} \ctext[RGB]{152,223,138}{\_} \ctext[RGB]{255,187,120}{status} \ctext[RGB]{152,223,138}{\_} \ctext[RGB]{255,187,120}{code}} \\
        \texttt{\ctext[RGB]{255,187,120}{document}\ctext[RGB]{152,223,138}{\_}\ctext[RGB]{255,187,120}{type}} & \texttt{\ctext[RGB]{255,187,120}{document} \ctext[RGB]{152,223,138}{\_} \ctext[RGB]{255,187,120}{type}} \\\midrule
    \multicolumn{2}{l}{\textit{Dot notation in column references (add space)}}\\
    \texttt{\ctext[RGB]{255,187,120}{farm}\ctext[RGB]{152,223,138}{.}\ctext[RGB]{255,187,120}{co}\ctext[RGB]{152,223,138}{w}\ctext[RGB]{255,187,120}{s}} & \texttt{\ctext[RGB]{255,187,120}{farm} \ctext[RGB]{152,223,138}{.} \ctext[RGB]{255,187,120}{cows}}\\
    \texttt{\ctext[RGB]{255,187,120}{origin}\ctext[RGB]{152,223,138}{.}\ctext[RGB]{255,187,120}{f}\ctext[RGB]{152,223,138}{light}} & \texttt{\ctext[RGB]{255,187,120}{origin} \ctext[RGB]{152,223,138}{.} \ctext[RGB]{255,187,120}{flight}}\\\midrule
    \multicolumn{2}{l}{\textit{SQL keyword (expand spelling)}}\\
    \texttt{\ctext[RGB]{255,187,120}{a}\ctext[RGB]{152,223,138}{v}\ctext[RGB]{255,187,120}{g}} & \texttt{\ctext[RGB]{255,187,120}{average}}\\
    \texttt{\ctext[RGB]{255,187,120}{des}\ctext[RGB]{152,223,138}{c}} & \texttt{\ctext[RGB]{255,187,120}{descending}}\\\bottomrule
    \end{tabular}
    }
    \caption{Three token preprocessing types. Coloring indicates tokenization, same as Table~\ref{tab:overview}.}
    \label{tab:token-preprocessing}
\end{table}
We present our two techniques for improving the generalization of LM-based semantic parsers.
LM pre-training learns high-quality contextualized word representation \citep{devlin-etal-2019-bert}, but to effectively use it on a downstream task, the tokenization needs to ``make sense.'' For example, if the text ``pet\_age'' is tokenized as ``pet'', ``\_'' and ``age'', then the semantics of ``pet'' and ``age'' acquired during pretraining can be directly used. However, if it is tokenized as ``pe'', ``t\_a'' and ``ge'', then pre-training is hardly useful because the model does not even recognize the two semantic words. 

Unfortunately, this latter case is very common when tokenizing non-natural language texts, such as database schemas and SQL queries. Thus, we propose a token preprocessing method to induce more natural tokenization by, at a high level, adding white spaces and handling the naming conventions in database schema and SQL queries. We show examples in Table~\ref{tab:token-preprocessing} and details in Appendix~\ref{app:token}.

\subsection{Component Boundary Marking} \label{sssec:sequence_augmentation}
At the sequence level, our second technique further assists LMs in recognizing the semantic boundaries of components aligned between input and output. An example is shown in Table~\ref{tab:overview}. While prior work has attempted the goal via implementing alignment-based attention supervision \cite{yin-etal-2021-compositional}, we propose to insert \emph{special tokens} in input and output to inject such bias. Specifically, we use pairs of ``\texttt{[sep$N$]}'' and ``\texttt{[/sep$N$]}'', $N \in \mathbb{Z}$, to mark the boundaries, so as to hint the LM that components within the paired special tokens should be aligned. 

In practice, we also observed cases where an NL component has to be aligned with a SQL component consisting of multiple non-continuous segments. To handle it, we will apply the same pair of special tokens to each segment of the same component. An example is shown in Table~\ref{tab:comp-boundary-marking-examples} in the Appendix.

Finally, we note that our method assumes the availability of component annotations. Such annotations can be obtained via human labeling \cite{gan-etal-2021-natural-sql}, heuristic rules \cite{yin-etal-2021-compositional}, or other advanced machine learning algorithms, but this is beyond the scope of our work.

\section{Experiments}

\subsection{Setup}
\begin{table}[t!]
    \centering
    \resizebox{0.9\columnwidth}{!}{%
    \begin{tabular}{lrll}
    \toprule
    \textbf{Dataset} & \textbf{Size} & \textbf{Usage} & \textbf{Generalization Type} \\
    \midrule
    Spider$_T$ & 7,000 & Train & None (in-distribution) \\
    Spider$_D$ & 1,034 & Eval & Domain \\
    CG-SUB$_T$ & 20,686 & Eval & None (in-distribution) \\
    CG-SUB$_D$ & 2,883 & Eval & Domain \\
    CG-APP$_T$ & 18,793 & Eval & Composition \\
    CG-APP$_D$ & 3,237 & Eval & Domain \& Composition \\
    \bottomrule
    \end{tabular}
    }
    \caption{Datasets in our experiments.}
    \label{tab:eval_sets}
\end{table}

\paragraph{Datasets.}
We use two datasets, Spider \citep{yu-etal-2018-spider} and Spider-CG \citep{gan-etal-2022-measuring}. Spider consists of a training set (\spidert{}) and a development set (\spiderd{}) with non-overlapping domains but otherwise similar data characteristics (e.g., length). Thus, we train the models on \spidert{}, and consider \spiderd{} as the evaluation for domain generalization. Spider-CG is derived from Spider by first dissecting each Spider instance into different components according to its dependency parse and generates data in two ways: substituting a component in one instance with one from another instance and appending one component from one instance to another instance. Depending on whether the instances come from the Spider training or development set, we get four splits: \cgsubt{}, \cgsubd{}, \cgappt{} and \cgappd{}, all of which are only used for evaluation. The instances created under substitution share similar data characteristics while those under appending are much longer, so a good model performance on the latter requires compositional generalization. Table \ref{tab:eval_sets} summarizes the dataset information. In addition, we use the NatSQL representation \citep{gan-etal-2021-natural-sql} throughout the experiment due to its better alignment with the NL input. 

\begin{table*}[t!]
    \centering
    \resizebox{0.95\textwidth}{!}{%
    \begin{tabular}{lcccccccccc}
    \toprule
    \multirow{2}{*}{\textbf{Model}} & \multicolumn{2}{c}{\textbf{Spider$_D$}} & \multicolumn{2}{c}{\textbf{CG-SUB$_T$}} & \multicolumn{2}{c}{\textbf{CG-SUB$_D$}} & \multicolumn{2}{c}{\textbf{CG-APP$_T$}} & \multicolumn{2}{c}{\textbf{CG-APP$_D$}}\\
    & \textbf{EM} & \textbf{EX} & \textbf{EM} & \textbf{EX} & \textbf{EM} & \textbf{EX} & \textbf{EM} & \textbf{EX} & \textbf{EM} & \textbf{EX}\\
    \toprule
    \multicolumn{11}{l}{\textit{Semantic Parsers with Specialized Architectures \cite{gan-etal-2022-measuring}}}\\ 
    RATSQL$_{B(S)}$ & 71.9 & - & 91.0 & - & 72.6 & - & 79.8 & - & 61.5 & - \\
    RATSQL$_{G(S)}$ & \bftab 74.5 & - & \bftab 91.4 & -  & \bftab76.7 &- & \bftab82.5 & - & \bftab68.3 & - \\
    \midrule
    \multicolumn{11}{l}{\textit{Semantic Parsers based on LMs}}\\ 
    T5-base & 64.6 & 67.9 & 83.8 & 88.1  & 69.1 & 71.1 & 60.2 & 70.3 & 45.0 & 54.9 \\
    T5-base + Tok &  \ul{71.8} &  \ul{75.6} & 85.9 & 89.5  & 74.1 & 78.6 & 65.2 & 73.8 & 54.2 & 65.9  \\
    T5-base + Comp & 64.4 & 68.2 & 86.3 & 90.2  & 69.3 & 73.1 & 69.8 &  \ul{77.9} & 53.5 & 63.4 \\
    T5-base + Tok + Comp & 69.4 & 73.2 &  \ul{86.6} & \ul{90.7}  &  \ul{76.6} & \ul{79.8} &  \ul{71.1} & 77.8 &  \ul{61.0} &  \ul{69.4}\\
    T5-base + Tok + Attn. Sup & 69.4 & 73.7 & 83.6 & 87.7  & 71.7 & 75.6 & 62.3 & 70.8 & 56.3 & 66.2\\
    \bottomrule
    \end{tabular}
    }
    \caption{Results (\%) on different evaluation sets. 
    Top: state-of-the-art model using specialized architecture; numbers are collected from its paper and only EM is reported (code unavailable).
    Bottom: T5-base models with our proposed or baseline techniques; we report the average performance of each model over three runs. \textbf{Tok}: token preprocessing. \textbf{Comp}: component boundary marking. \textbf{Attn. Sup}: the attention supervision method of \citet{yin-etal-2021-compositional}.}
    \label{tab:results}
\end{table*}

\noindent \textbf{Evaluation Metrics.} We follow the standard Spider benchmarking and employ two evaluation metrics. \textbf{Exact Match (EM)} compares the generated and the ground-truth query by performing exact set matching at the lexical level \cite{yu-etal-2018-spider}. \textbf{Execution Match (EX)} measures whether executing the generated query on the given database can yield the same results as using the ground truth. Notably, for a fair comparison with existing semantic parsers on the Spider leader board, we follow \citet{gan-etal-2022-measuring}, convert each generated NatSQL query into a SQL query, and report the evaluation results based on the converted SQL query.

\noindent \textbf{Models, Baselines, and Implementation.}
We evaluate our proposed techniques by applying them to the pre-trained T5 model \cite{raffel2020exploring}. Our experiments are conducted using T5-base, with the use of database contents following \citet{lin-etal-2020-bridging}.
As our second technique leverages component boundary labels to encourage the compositional generalization of LM, we compare it with a baseline \cite{yin-etal-2021-compositional} which similarly assumes the labels but utilizes them in a more complicated way, i.e., transforming the component alignments into supervision on the cross attention between input and output of the LM. We denote this baseline as \textbf{Attn. Sup}.\footnote{In our implementation, we apply the supervision to cross-attention distribution averaged across all decoder layers and heads. We also tried cross-attention from only the top decoder layer, but the results are similar.}
For both methods, we leverage component annotations from Spider-SS \cite{gan-etal-2022-measuring}.
 
These annotations were generated by applying a syntactic parser to decompose the NL question into sub-questions and then manually annotating their corresponding NatSQL components. 

We also compare with the state-of-the-art models, RATSQL$_{B(S)}$ and RATSQL$_{G(S)}$, from \citet{gan-etal-2022-measuring}, although their models adopt a specialized architecture (i.e., RATSQL \cite{wang-etal-2020-rat}) and RATSQL$_{G(S)}$ additionally employed task-specific pre-training \cite{shi2021learning}. Both models used the same component annotations from Spider-SS.

Finally, for each of our model variants in Table~\ref{tab:results}, we repeat the experiment three times, using three random seeds consistently across all models, and report the average results.
We include more implementation details in Appendix~\ref{app:implement}.


\subsection{Results}

\paragraph{Main Results.}
We present our results in Table \ref{tab:results}. First, all models obtain the best performance on the in-distribution evaluation set CG-SUB$_T$ while suffering from more than 10\% performance drops on others, confirming the challenges of the domain and compositional generation. As expected, all models have the worst performance on CG-APP$_D$, which requires both types of generalization. Between the two types, it is also observed that compositional generalization (as measured by CG-APP$_T$) is more challenging than domain generalization (as measured by Spider$_D$ and CG-SUB$_D$).

Second, our results show that the token preprocessing method, albeit simple, can improve both domain and compositional generalizations of LMs dramatically. For example, comparing T5-base with T5-base+Tok, the latter is improved by around 5-7\% EM and 7\% EX for domain generalization (on Spider$_D$ and CG-SUB$_D$), 5\% EM and 3.5\% EX for compositional generalization (on CG-SUB$_T$), and 9\% EM and 11\% EX for the challenging case when both types occur (on CG-APP$_D$). {Additionally, we also show the effectiveness of token preprocessing with T5-3B on Spider$_D$ in App. \ref{t5-3b}.}

Moving on to our proposed component boundary marking method, it shows to be particularly helpful for compositional generalization.
Specifically, applying it to T5-base leads to a 9\% EM and 7\% EX increase on CG-APP$_T$, and an 8\% EM and 8\% EX increase on CG-APP$_D$. On the in-distribution evaluation set, this technique also gives slight improvement, whereas, for domain generalization, there is no obvious impact from this technique.

Finally, augmenting T5-base with both techniques (i.e., T5-base+Tok+Comp) leads to better performance than applying each technique individually in most evaluation sets, implying that our two techniques are complementary to each other. Specifically, for in-distribution evaluation, using each technique individually or both of them together yield similar results; for domain generalization, there is no additional gain from applying component boundary marking on the top of the token preprocessing; for compositional generalization, the two techniques together contribute the best EM across all models and baselines. Overall, combining the two techniques shrinks the performance gap between in-distribution and domain OOD by around 2-4\% EM, composition OOD by 7\%, and joint OOD by 13\%.

\noindent \textbf{Compared with Special Architectures.} Despite its simplicity, our T5-base+Tok+Comp model achieves comparable or better performance than the two RATSQL variants on CG-SUB$_D$. It also performs comparably to RATSQL$_{B(S)}$ on CG-APP$_D$.

\vspace{1mm}
\noindent \textbf{Compared with Attn. Sup.}
Surprisingly, the attention supervision has only led to around 2\% EM and 1.5\% EX gains on CG-APP$_D$, while no further advantage is observed on other evaluation sets. In our conjecture, this is due to the misalignment between the objective of Attn. Sup \cite{yin-etal-2021-compositional} and the attention mechanism of pre-trained LMs. Specifically, Attn. Sup encourages the attention distribution of different heads to be consistent with the component alignment supervision. However, prior work \cite{voita-etal-2019-analyzing} suggests that different attention heads of even the same layer may have different functions and roles. Thus, when coarsely defining the objective function, it may not allow for the most effective supervision. Furthermore, similar to our finding, \citet{yin-etal-2021-compositional} did not observe performance gain when they applied Attn. Sup to T5-base on CFQ \cite{keysers2020measuring}.

\vspace{1mm}
\noindent \textbf{Qualitative Analysis on Tokenization.}
To qualitatively understand how our token preprocessing helps the generalization, we randomly sampled 50 examples from the Spider$_D$ to analyze how frequently the T5 tokenizer divides tokens into less meaningful subtokens. Consequently, we found 243 tokenization issues in total, and 140 of them can be resolved by our token preprocessing. The remaining cases are like splitting ``id'' into ``i'' and ``d'' as shown in Table~\ref{tab:overview}, which is beyond our scope.

\vspace{1mm}
\noindent \textbf{Error Analysis on Component Boundary Marking.}
 We manually examined 50 error predictions from T5-base+Tok+Comp and contrasted them with the errors of T5-base+Tok. Intriguingly, we observed much more frequent schema items or value hallucinations from the former. For example, it may generate queries accessing non-existing columns in a table, or misspells the literal values in the queries. We conjecture that this is because our component boundaries are only applied to the NL input, not the database schema (note that literal values are grounded and attached to schema items in their input representations; see Appendix~\ref{app:implement} for details). This reveals a new challenge of LM generalization in text-to-SQL semantic parsing, i.e., how to properly handle the database schema when injecting prior knowledge into LMs for compositional generalization.

\vspace{1mm}
\noindent \textbf{Generalizing to Other Semantic Parsing Tasks.}
While our main focus in this work is on text-to-SQL parsing, we also investigate whether our approaches can generalize beyond this specific task. To this end, we implemented both of our techniques to SMCalFlow-CS \cite{yin-etal-2021-compositional}, a compositional generalization dataset for text-to-LISP expression parsing \cite{SMDataflow2020}. For ``+Comp'', We utilize the span-level alignments heuristically derived by \citet{yin-etal-2021-compositional} as component annotations.\footnote{Yin et al.'s approach requires knowing the ground-truth LISP expression when deriving the component boundaries for the input question. In our experiment, we assume the availability of these question boundaries at test time and focus on showcasing the potential of ``Comp'', while automating this question decomposition is left as future work.} Our results in Table~\ref{tab:smcalflow} show that: 
(1) Our token preprocessing can be universally helpful for LMs to model schema items, predicates, etc., leading to 1.2\% performance gain over T5-base;
(2) Our component boundary marking method is highly effective for compositional generalization, which offers 2.6\% additional gain.
\begin{table}[t!]
    \centering
    \resizebox{0.9\columnwidth}{!}{%
    \begin{tabular}{lrll}
    \toprule
    \textbf{Model} & \textbf{Exact Match}  \\
    \midrule
    COARSE2FINE + SS (Span-level Sup.) & 47.4  \\
    \midrule
    T5-base & 63.9  \\
    T5-base + Tok & 65.1 \\
    T5-base + Tok + Comp & \bftab \ul{67.7} \\
    \bottomrule
    \end{tabular}
    }
    \caption{Results (\%) on SMCalFlow-Compositional Skills dataset (16-shot setting). Top: Result from \citet{yin-etal-2021-compositional}. Bottom: T5-base models with our proposed or baseline techniques; we report the average performance of each model over three runs.}
    \label{tab:smcalflow}
\end{table}

\section{Conclusion} 
In this paper, we present two simple yet effective techniques to improve the domain and compositional generalization of LMs in text-to-SQL semantic parsing. Our techniques aid LMs in preserving the semantic boundaries of tokens and components in their input and output. We also demonstrate their potential to be generalized to other semantic parsing tasks.

\section*{Limitations}
Future work can further apply our approaches to other semantic parsing tasks.
For example, for parsing texts to lambda-calculus expressions for knowledge base question answering \cite{dong-lapata-2016-language}, one can similarly preprocess the schema items (e.g., ``\texttt{department\_time}'' into ``\texttt{department \_ time}'') and typed values (e.g., ``\texttt{dallas:ci}'' into ``\texttt{dallas : ci}'') for more meaningful subword tokenization results.
In addition, our experiments are based on T5. To further verify the effectiveness of our techniques, one can apply them to other pre-trained language models such as BART \cite{lewis-etal-2020-bart} and GPT-2 \cite{radford2019language} as well.

\section*{Acknowledgments} 
We would like to thank all anonymous reviewers for their constructive comments. We also thank Yujian Gan and Xinyun Chen for their help in using the NatSQL and the Spider-SS datasets, as well as Pengcheng Yin for using the code base of Attn. Sup.
This project was supported by resources provided by the Office of Research Computing at George Mason University (\url{https://orc.gmu.edu}) and funded in part by grants from the National Science Foundation (Awards Number 1625039 and 2018631).

\bibliography{acl2023}
\bibliographystyle{acl2023}



\appendix

\section{Token Preprocessing Details}\label{app:token}
We propose a simple token preprocessing method. Instead of directly feeding the input to the subword tokenizer, we introduce three preprocessing steps: (1) For schema items in input and output, reversing the snake case to the normal, e.g., ``\texttt{pet\_age}'' to ``\texttt{pet \_ age}''; (2) For any call of ``\texttt{Table.Column}'', splitting the tokens around the access operator ``\texttt{.}'' (i.e., ``\texttt{Table . Column}'');  and (3) Replacing any reserved words that cannot be properly handled in NatSQL, e.g., ``\texttt{avg}'' to ``\texttt{average}''. In practice, we also handle formalism-specific special tokens, e.g., adding the ``less than'' operator ``\texttt{<}'' to the vocabulary of T5 tokenizer. While we showcase our token preprocessing under text-to-SQL parsing, the intuition can be generalized to other formalisms (e.g., regex, $\lambda$-expression) easily.  

In addition, we also check the issue of tokenization in other popular LM tokenizers and found that the tokenization issue is not specific to T5. Examples of bad tokenization from BERT \citep{devlin-etal-2019-bert} and GPT2 \citep{radford2019language} tokenizers and after our token preprocessing are listed in Table~\ref{tab:tokenizer}.

\begin{table}[h!]
    \centering
    \resizebox{0.9\columnwidth}{!}{
    \begin{tabular}{ll}\toprule
    \multicolumn{2}{l}{\textit{GPT2 Tokenizer}} \\
        \textbf{Before}: & \texttt{\ctext[RGB]{255,187,120}{student\_}\ctext[RGB]{152,223,138}{en}\ctext[RGB]{255,187,120}{rol}\ctext[RGB]{152,223,138}{ment}\ctext[RGB]{255,187,120}{\_}\ctext[RGB]{152,223,138}{c}\ctext[RGB]{255,187,120}{ourses}}\\
        \textbf{After}: & \texttt{\ctext[RGB]{255,187,120}{student} \ctext[RGB]{152,223,138}{\_} \ctext[RGB]{255,187,120}{enrolment} \ctext[RGB]{152,223,138}{\_} \ctext[RGB]{255,187,120}{courses}}\\
        \textbf{Before}: & \texttt{\ctext[RGB]{255,187,120}{trans}\ctext[RGB]{152,223,138}{cript}\ctext[RGB]{255,187,120}{s}\ctext[RGB]{152,223,138}{.}\ctext[RGB]{255,187,120}{trans}\ctext[RGB]{152,223,138}{cript}\ctext[RGB]{255,187,120}{\_}\ctext[RGB]{152,223,138}{date}}\\
        \textbf{After}: & \texttt{\ctext[RGB]{255,187,120}{transcript}\ctext[RGB]{152,223,138}{s} \ctext[RGB]{255,187,120}{.} \ctext[RGB]{152,223,138}{transcript} \ctext[RGB]{255,187,120}{\_} \ctext[RGB]{152,223,138}{date}}\\
        \textbf{Before}: & \texttt{\ctext[RGB]{255,187,120}{av}\ctext[RGB]{152,223,138}{g}}\\
        \textbf{After}: & \texttt{\ctext[RGB]{255,187,120}{average}}\\
        
        \midrule
    
        \multicolumn{2}{l}{\textit{BERT Tokenizer}} \\
        \textbf{Before}: & \texttt{\ctext[RGB]{255,187,120}{singer}\ctext[RGB]{255,187,120}{.}\ctext[RGB]{152,223,138}{Net}\ctext[RGB]{255,187,120}{Worth}\ctext[RGB]{152,223,138}{Mil}\ctext[RGB]{255,187,120}{lions}}\\
        \textbf{After}: & \texttt{\ctext[RGB]{255,187,120}{singer} \ctext[RGB]{152,223,138}{.} \ctext[RGB]{255,187,120}{Net} \ctext[RGB]{152,223,138}{Worth} \ctext[RGB]{255,187,120}{Millions}}\\
        \textbf{Before}: & \texttt{\ctext[RGB]{255,187,120}{av}\ctext[RGB]{152,223,138}{g}}\\
        \textbf{After}: & \texttt{\ctext[RGB]{255,187,120}{average}}\\
        \textbf{Before}: & \texttt{\ctext[RGB]{255,187,120}{as}\ctext[RGB]{152,223,138}{c}}\\
        \textbf{After}: & \texttt{\ctext[RGB]{255,187,120}{ascending}}\\
        \bottomrule
    \end{tabular}
    }
    \caption{Tokenization of snake case, camel case, and token notation in BERT and GPT2 tokenizer. Coloring indicates tokenization, same as Table \ref{tab:overview}.}
    \label{tab:tokenizer}
\end{table}

\section{T5-3B Experiment} \label{t5-3b}

\begin{table}[ht!]
    \centering
    \resizebox{0.9\columnwidth}{!}{%
    \begin{tabular}{lrll}
    \toprule
    \multirow{2}{*}{\textbf{Model}} & \multicolumn{2}{c}{\textbf{Spider$_D$}} \\
    & \textbf{EM} & \textbf{EX} \\
    \midrule
    T5-3B (w deepspeed) & 73.2  & 77.4 \\
    T5-3B (w/o deepspeed) & 76.0  & 79.8 \\
    T5-3B + Tok (w deepspeed) & 74.4  & 78.7 \\
    T5-3B + Tok (w/o deepspeed) & \bftab \ul{77.4} & \bftab \ul{80.9} \\
    \bottomrule
    \end{tabular}
    }
    \caption{Results (\%) on Spider$_D$ when T5-3B(+Tok) was trained with or without using deepspeed. }
    \label{tab:t5-3b}
\end{table}
{To assess the effectiveness of our token preprocessing technique with larger LMs, we apply it to T5-3B and evaluate the model on Spider$_D$. The results are shown in Table \ref{tab:t5-3b}. Our results show that T5-3B+Tok has a performance gain of 1.1\%, indicating that it is helpful for larger LMs as well. Additionally, we also provide results with and without using \citet{github_deepspeed}, a deep learning optimization library that is used to train large models more efficiently. Surprisingly, although \citet{github_deepspeed} helped us improve training speed, we found a performance drop of around 2.1-2.2\% EX while using it. However, our token preprocessing consistently leads to around 1.0\% absolute performance gain.}

\section{Component Boundary Marking Details}

In Table~\ref{tab:comp-boundary-marking-examples}, we present one more example of component boundary marking. In this example, the NL component \textit{``What is the most populace city''} is aligned with two non-continuous SQL segments, \texttt{``select city.Name, city.Population''} and \texttt{``order by city.Population desc limit 1''}. To handle such cases, we apply the same pair of special tokens \texttt{``[sep0]'' ``[/sep0]''} twice, one for each segment.

\begin{table}[h!]
    \centering
    \resizebox{\columnwidth}{!}{
    \begin{tabular}{p{1cm}p{9cm}}\toprule
        
        \multicolumn{2}{p{10cm}}{\textit{Component Boundary Marking Example}}\\\midrule
        \textbf{Before}: & What is the most populace city that speaks English?  \\
        & \texttt{Select city.Name, city.Population where countrylanguage.Language = “English” order by city.Population desc limit 1 }\\
        \textbf{After}: & \texttt{\textcolor[RGB]{255,127,14}{[sep0]}} What is the most populace city \texttt{\textcolor[RGB]{255,127,14}{[/sep0]}} \texttt{\textcolor[RGB]{255,127,14}{[sep1]}} that speaks English? \texttt{\textcolor[RGB]{255,127,14}{[/sep1]}}\\
        & \texttt{\textcolor[RGB]{255,127,14}{[sep0]}  select city.Name , city.Population \textcolor[RGB]{255,127,14}{[/sep0]} \textcolor[RGB]{255,127,14}{[sep1]} where countrylanguage.Language = "English" \textcolor[RGB]{255,127,14}{[/sep1]} \textcolor[RGB]{255,127,14}{[sep0]} order by city.Population desc limit 1 \textcolor[RGB]{255,127,14}{[/sep0]}} \\
    \bottomrule
    \end{tabular}
    }
    \caption{An example of component boundary marking when an NL component aligns with non-continuous segments in the SQL side. In this case, we apply the special tokens for each segment.}
    \label{tab:comp-boundary-marking-examples}
\end{table}

\section{Implementation Details}\label{app:implement}
Our experiments are conducted based on the pre-trained T5 model. The input to T5 follows the same format and order as \citet{scholak-etal-2021-picard} (except our additional token preprocessing, if applied), i.e., ``\texttt{Question | Database 1 | Table 1: Column 1, Column 2,...| Table 2: Column 1, Column 2...}''. We also use the database contents as parts of the input, following \citet{lin-etal-2020-bridging}. For example, if the NL question mentions a literal value (e.g., ``New York''), the appearance of whom can be found in the contents of a certain ``\texttt{Column 1}'' via fuzzy string matching, then when we represent the database schema, we will include it via ``\texttt{Database 1 | Table 1: Column 1 (New York), Column 2, ...}''.  

We fine-tune the T5-base LM that consists of 220 million parameters on NVIDIA A100 GPU for 10-12 hours. It was trained with a learning rate of $10^{-4}$ and batch size 16 for T5-base for a maximum of 20K training steps. The model is evaluated on Spider$_D$ for every 1K training steps, and the best checkpoint is selected based on the model EM on Spider$_D$. In inference time, we perform simple greedy decoding. 

We use the PyTorch-Transformers library \citep{wolf-etal-2020-transformers}, which is a library for state-of-the-art pre-trained models for NLP, to fine-tune our models. Specifically, our code for fine-tuning T5-base is adapted from PICARD's implementation \citep{scholak-etal-2021-picard}. {Furthermore, we also use \citet{github_deepspeed} to fine-tune all of our T5-base models.}

\paragraph{Datasets.} We used Spider \cite{yu-etal-2018-spider}, NatSQL \cite{gan-etal-2021-natural-sql}, Spider-CG \cite{gan-etal-2022-measuring}, and SMCalFlow-CS \cite{yin-etal-2021-compositional} in our work. They are under the license of CC BY-SA 4.0. Our use of these datasets is consistent with their intended use, i.e., for scientific research.
All datasets are in English. They contain annotated NL and SQL or NatSQL or LISP expression pairs from the open domain.

\end{document}